\documentclass[times, 10pt,twocolumn]{article}
\usepackage{latex12}
\usepackage{times}
\usepackage{balance}
\usepackage{epsfig}
\usepackage{graphicx}
\usepackage{amsmath}
\usepackage{subfig}
\usepackage{siunitx}
\begin{document}

\title{Tracking \textit{Tetrahymena Pyriformis} Cells using Decision Trees}

\author{Quan Wang, Yan Ou, A. Agung Julius, Kim L. Boyer\\
\emph{Rensselaer Polytechnic Institute}\\
\emph{wangq10@rpi.edu}
\and
Min Jun Kim\\
\emph{Drexel University}\\
\emph{mkim@coe.drexel.edu}\\
}

\maketitle
\thispagestyle{empty}

\begin{abstract}
Matching cells over time has long been the 
most difficult step in cell tracking. 
In this paper, we approach this problem
by recasting it as a classification problem. 
We construct a feature set for each cell, 
and compute a feature difference vector between a cell
in the current frame and a cell in a previous frame.
Then we determine whether the two cells
represent the same cell over time by training decision trees
as our binary classifiers. With the output of decision trees, 
we are able to
formulate an assignment problem for our cell association task
and solve it using a modified version of
the Hungarian algorithm. 
\end{abstract}

\section{Introduction}
Researchers have been creating artificial magnetotactic 
\textit{Tetrahymena pyriformis} (\textit{T. pyriformis}) cells 
by the internalization of iron oxide
nano particles, and controlling them 
with a time-varying external magnetic field \cite{cell1,ouyan}. 
To perform the multi-cell control task, it is necessary
to track the real-time position of each cell and use it as the feedback for the control system. 
However, this problem is very difficult because: 
the \textit{T. pyriformis} cells are moving fast
in the field of view; several cells may overlap and occlude each
other; the appearances of different cells can be similar; and the same cell
may change in appearance over time \cite{tracking_cell}.  

Decision tree was proposed as a powerful machine learning technique, 
which can be used for both classification and regression, 
and has been successfully applied on practical systems such as the Kinect 
gaming platform \cite{kinect}. 
Since the decision tree can take very complicated 
features as its input and at the same time, once a decision tree is trained, the decision procedure can be very fast, it is ideal for real-time classification. 
We use decision trees as a classifier to determine whether
regions segmented in neighboring and non-neighboring frames 
represent the same \textit{T. pyriformis} cell. 

\section{Background Subtraction}
To extract the foreground regions of cells, we first perform a background subtraction at 
each frame. 
There are a number of well-known background subtraction algorithms such as adaptive Gaussian mixture models. 
Since the background in our video is simple and consistent over time, 
we simply apply a median filter on the
time dimension for the first 20 frames to obtain the background image. 
At each frame, we subtract the background image from
the current frame and take the absolute value to obtain a difference image. 
Then we apply thresholding, fill the morphological holes, 
and extract the connected regions 
as candidate \textit{T. pyriformis} cells.

\section{Feature Extraction}
\begin{figure}
  \centering
  \subfloat[]
  {\includegraphics[width=0.2\textwidth]{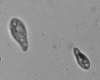} }  
  \;      
  \subfloat[]
  {\includegraphics[width=0.2\textwidth]{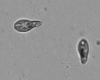} }
  \caption{Examples of the \textit{T. pyriformis} cell appearance. }
  \label{fig:cell_appearance}
\end{figure}

After the region of each cell is segmented, we extract
features for each cell according to their shapes, gray-level intensities, 
and the combination of both. 
Examples of the \textit{T. pyriformis} cell appearance in our video are
shown in Figure \ref{fig:cell_appearance} (pixel spacing = \SI{2.32/2.32}{\micro\metre}).
Assume a cell $C$ in one frame consists of $N$ pixels $I(\textbf{x}_i)$, where
$i=1,2,\dots,N$, and $\textbf{x}_i=(x_i,y_i)$ are the coordinates of the $i$th pixel
of $C$. We can define some useful features for this cell, as follows. 

\paragraph{Spatial Features}
The cell's area $N$ and centroid $\textbf{x}_c=(x_c,y_c)$, 
where $x_c=\overline{x_i}$ and $y_c=\overline{y_i}$,
are the most basic features. 
Another useful spatial feature is the
$n$th order normalized inertia \cite{inertia}, which measures the circularity 
of the shape, and is defined as
\begin{equation}
\label{eq:inertia}
J_n(C)=\dfrac{\sum\limits_{i=1}^N ||\textbf{x}_i-\textbf{x}_c ||^n}
{N^{1+n/2}} .
\end{equation}

\paragraph{Histogram Features}
Using only pixel intensities without spatial positions, 
we can compute several histogram features. 
First we have the mean and the standard deviation:
\begin{eqnarray}
\label{eq:mu}
\mu(C)&=&\dfrac{1}{N}\sum\limits_{i=1}^N I(\textbf{x}_i) ,
\\
\label{eq:sigma}
\sigma(C)&=&\dfrac{1}{N} 
\sqrt{\sum\limits_{i=1}^N \left( I(\textbf{x}_i)-\mu(C) \right)^2} .
\end{eqnarray}
Then we have the standard moments such as the skewness $\gamma(C)$
and the kurtosis $\beta(C)$:
\begin{eqnarray}
\label{eq:gamma}
\gamma(C)&=&E \left[ \left( 
\dfrac{I(\textbf{x}_i)-\mu(C)}{\sigma(C)} \right)^3 \right] ,
\\
\label{eq:beta}
\beta(C)&=&E \left[ \left( 
\dfrac{I(\textbf{x}_i)-\mu(C)}{\sigma(C)} \right)^4 \right] ,
\end{eqnarray}
where $E[\cdot]$ denotes expectation. 
We also use the $n$th order root of the $n$th order central moment $M_n(C)$:
\begin{eqnarray}
\label{eq:moment}
M_n(C)= \sqrt[n] { E \left[ \left(I(\textbf{x}_i)-\mu(C)\right)^n\right]  } .
\end{eqnarray}

\paragraph{Composite Features}
Finally we can compute some features that use both spatial information and pixel  intensities 
at the same time to capture the spatial distribution of pixel intensities. 
We first define
the polynomial distribution feature $P_n(C)$:
\begin{eqnarray}
\label{eq:polynomial}
P_n(C)=\dfrac{1}{N^{1+n/2}}\sum\limits_{i=1}^N ||\textbf{x}_i-\textbf{x}_c ||^n \cdot I(\textbf{x}_i).
\end{eqnarray}
Then we define the Gaussian distribution feature $G_n(C)$:
\begin{eqnarray}
\label{eq:gaussian}
G_n(C)=\dfrac{1}{N}\sum\limits_{i=1}^N 
\exp\left(-\dfrac{||\textbf{x}_i-\textbf{x}_c ||^2}{2n^2} \right) \cdot I(\textbf{x}_i).
\end{eqnarray}
These composite features can be viewed as weighted means of pixel intensities. 
For example, $P_n(C)$ gives larger weights to pixels far away from the center while
$G_n(C)$ gives larger weights to pixels near the center. 

\section{Decision Trees}
To use decision trees for the cell association problem, we first need to map the 
cell association problem to a classification problem. For each cell $L$ in a previous frame
 and each cell $C$ in the current frame, we need to know whether they are the same
cell. Thus the classification problem is 
to develop a binary classifier $f$
such that $f(C,L)=1$ if $C$ and $L$ are the same cell, and $f(C,L)=0$ if they are not. 

\subsection{Feature Difference Vector}
To simplify the input to the classifier, we define a 23-dimensional feature difference vector $\textbf{v}(C,L)=(v_1,v_2,\dots,v_{23})$ for $C$ and $L$. 
Each entry of this feature difference vector reflects the difference of the two cells in some aspect. 
$v_1$ is the distance between 
the center of $C$ and the 
predicted center of $L$ in the current frame. 
A simple constant velocity assumption is used for the prediction. 
For example, suppose the current frame is the $k$th frame and the center of $C$ is $\textbf{x}_c(C)$, the center of $L$ at the $(k-1)$th frame was $\textbf{x}_c(L)$ and at the $(k-2)$th frame was $\textbf{x}'_c(L)$. 
Then the predicted center
of $L$ at the current frame is $\textbf{x}_p(L)=2\textbf{x}_c(L)-\textbf{x}'_c(L)$, and
$v_1$ is defined as
\begin{equation}
\label{eq:predicted_distance}
v_1=|| \textbf{x}_c(C) - \textbf{x}_p(L) || .
\end{equation}
$v_2$ is defined as the distance between the centers of the two cells:
\begin{equation}
v_2=|| \textbf{x}_c(C) - \textbf{x}_c(L) || .
\end{equation}
Each of the other 21 entries of $\textbf{v}(C,L)$ is defined as the absolute value of the 
difference of some feature of $C$ and $L$. For example, let $N(C)$ and $N(L)$ be the areas 
of $C$ and $L$, respectively. Then we have
\begin{eqnarray}
v_3&=&| N(C)-N(L)| , \\
v_4&=&| \mu(C)-\mu(L) | .
\end{eqnarray}
Similarly, $v_5$ to $v_7$ are defined using Equations (\ref{eq:sigma}) to (\ref{eq:beta}) respectively; 
$v_8$ to $v_{11}$ are defined using (\ref{eq:moment}) for $n=3,4,5,6$; 
$v_{12}$ to $v_{15}$ are defined using (\ref{eq:inertia}) for $n=1,2,3,0.5$; 
$v_{16}$ to $v_{19}$ are defined using (\ref{eq:polynomial}) for $n=1,2,3,0.5$; and
$v_{20}$ to $v_{23}$ are defined using (\ref{eq:gaussian}) for $n=2,4,6,8$.
With these definitions, our classifier $f$ can be viewed as a mapping from the 23-dimensional space to the $\lbrace 0,1 \rbrace$ set. Thus we denote it as $f(\textbf{v}(C,L))$. 
 
\subsection{Training a Decision Tree}
\begin{figure}
  \centering
  {\label{fig:diagonal}
  \includegraphics[width=0.3\textwidth]{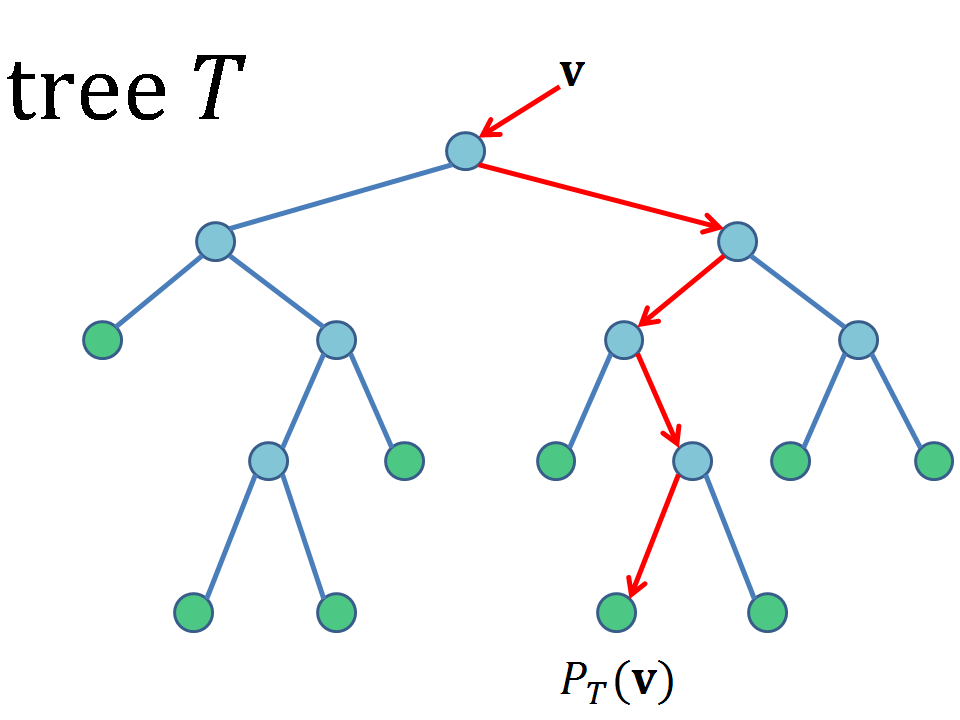} }  
  \caption{Decision tree example. A decision tree $T$ consists of 
  split nodes (blue) and leaf nodes (green). The red arrows show the 
  branching path for an input $\textbf{v}$.}
  \label{fig:tree}
\end{figure}

As shown in Figure \ref{fig:tree}, a decision tree $T$ consists of leaf nodes and split (non-leaf) nodes. Each split node consists of a threshold $\tau$ and an indicator $k$ for a feature $v_k$, where $k=1,2,\dots,23$. To classify the feature difference vector $\textbf{v}$, one starts from the root and repeatedly compares its $k$th entry $v_k$ with the threshold $\tau$ 
to decide whether to branch to the left sub-tree (if $v_k\leq \tau$) 
or right sub-tree (if $v_k> \tau$). 
The leaf node is a quantity $P_T(\textbf{v})$ 
indicating the probability of $f(\textbf{v}(C,L))=1$. 

Since it is easy for a human to distinguish whether two regions in different frames 
represent the same cell, we can manually assign labels $\lbrace y \rbrace$ to cell pairs 
from neighboring frames in training videos and associate these labels with extracted feature difference vectors $\lbrace \textbf{v} \rbrace$. In this way we build our training data set 
$D=\lbrace (\textbf{v},y ) \rbrace$, where $y=f(\textbf{v})$. 

The training process follows a maximal entropy reduction procedure \cite{kinect}:
\begin{enumerate}
\item Beginning from the root, consider a large set of splitting candidates $(k,\tau)$
which covers all possible $k$ and provides sufficiently delicate subdivisions for
each $v_k$. 
\item For each candidate $(k,\tau)$, we partition the training set $D=\lbrace (\textbf{v},y ) \rbrace$ into
two subsets:
\begin{eqnarray}
D_l (k,\tau) & = & \lbrace (\textbf{v},y) | v_k \leq \tau  \rbrace , \\
D_r (k,\tau) & = & D \setminus D_l(k,\tau) .
\end{eqnarray}
\item Find the candidate $(k,\tau)$ that maximizes the entropy reduction $G(k,\tau)$:
\begin{eqnarray}
(k^*,\tau^*)=\underset{(k,\tau)}{\operatorname{argmax}} \; G(k,\tau)  , 
\end{eqnarray}
\begin{eqnarray}
G(k,\tau)&=&H(D)
-\dfrac{|D_l(k,\tau)|}{|D|} H(D_l(k,\tau)) \nonumber \\
&&-\dfrac{|D_r(k,\tau)|}{|D|} H(D_r(k,\tau)) ,
\end{eqnarray}
where $H(\cdot)$ denotes the Shannon entropy and $|\cdot|$ denotes 
the cardinality of a set.
\item Use $(k^*,\tau^*)$ as the feature indicator and threshold at the current split node, 
and repeat the above steps for the left sub-tree with $D_l(k^*,\tau^*)$ and right sub-tree with $D_r(k^*,\tau^*)$. 
\item If the depth reaches the maximum, or the size (or entropy) of the partitioned subset 
$\widetilde{D}$ at the current node is sufficiently small, then we set 
this node as a leaf node, and the probability at this node is
\begin{eqnarray}
P_T=\dfrac{|\lbrace (\textbf{v},y)\in \widetilde{D} | y=1 \rbrace|}{|\widetilde{D}|} .
\end{eqnarray}
\end{enumerate} 
In our work, a tree $T_1$ is trained using $D=\lbrace {(\textbf{v},y)} \rbrace$, 
and another tree $T_2$ is trained using $D'=\lbrace {(\textbf{v}',y)} \rbrace$, where
$\textbf{v}'=(v_3,v_4,\dots,v_{23})$ is the truncated feature difference vector. 
Since $T_2$ is trained without center position information, we
use $T_1$ for cell association in neighboring frames and
$T_2$ for cell association in non-neighboring frames.

\section{Association Rules}
To track cells over time, we maintain a list $\lbrace L_j \rbrace$ 
of cells that have appeared in previous frames. The cell association task
is to determine which cell $L_j$ should be associated with the cell $C_i$ in the current frame. Each cell in the list has a status, being \textit{active}, \textit{occluded}, 
\textit{out}, or \textit{new}. 

\subsection{Association Matrix}
At frame $k$, if we have $N_1$ cells $\lbrace C_i \rbrace$ in the current frame 
and $N_2$ cells $\lbrace L_j \rbrace$ in the list, we define an $N_1\times(N_2+2)$ association matrix $A$. 
Each entry of $A$ represents the confidence of associating $C_i$
with a cell in the list or considering it as a new cell. 

If $1 \leq j \leq N_2$ and the status of $L_j$ is \textit{active}, \textit{new} or \textit{occluded}, 
we define 
\begin{equation}
A_{i,j}=P_{T_1}(\textbf{v}(C_i,L_j)) . 
\end{equation}

Let $d_0$ denote the maximum possible speed of 
a \textit{T. pyriformis}  cell in pixels per frame, and
$d_b(C_i)$ denote the distance from $C_i$ to the image border. 
A new cell or a re-appearing cell must (re-)enter the image from the image border. 
Thus for $1 \leq j \leq N_2$, if the status of $L_j$ is \textit{out}
and $d_b(C_i) > d_0$, 
we set $A_{i,j}=-1$; if the status of $L_j$ is \textit{out} and $d_b(C_i) \leq d_0$, 
we define 
\begin{equation}
A_{i,j}=\alpha_1 P_{T_2}(\textbf{v}'(C_i,L_j)) , 
\end{equation}
where $0<\alpha_1<1$ is a constant denoting the 
diffidence (as shown in Table \ref{table:tree_error}) of associating cells in non-neighboring frames.

$A_{i,N_2+1}$ is the confidence of considering $C_i$ as a new cell that has not appeared
in previous frames. If $d_b(C_i) > d_0$, we set $A_{i,N_2+1}=-1$;
if $d_b(C_i) \leq d_0$, we define 
\begin{equation}
A_{i,N_2+1}=\alpha_2 \exp ( -\lambda_1 d_b^2(C_i) ) , 
\end{equation}
where $0<\alpha_2<1$ and $\lambda_1>0$ are two constants. 

To also include the case when one cell is occluded by another, we use $A_{i,N_2+2}$
to represent the confidence that $C_i$ is a region of more than one cell 
overlapping. Let $\lbrace L_j' \rbrace $ be a subset of the list containing only cells that 
have appeared in the $(k-1)$th frame. We define 
\begin{equation}
\label{eq:N2+2}
A_{i,N_2+2}=\alpha_3
\sum\limits_{L_j'} e^{-\lambda_2 d_p^2(C_i,L_j')}
-\alpha_3 \sum\limits_{C_j} e^{-\lambda_2 d_c^2(C_i,C_j)} , 
\end{equation}
where $0<\alpha_3<1$ and $\lambda_2>0$ are two constants, $d_c(C_i,C_j)$ is the center distance between
$C_i$ and $C_j$, and 
$d_p(C_i,L_j')$ is the 
distance from the center of $C_i$ to the predicted center of $L_j'$, 
which is similar to (\ref{eq:predicted_distance}). 
The right-hand side of (\ref{eq:N2+2}) can be thought of as the predicted number of cells minus the 
observed number of cells near $C_i$. 

\subsection{Association List}
Once the association matrix $A$ is created, we need to associate each $C_i$ 
(or each row of $A$) with 
a column of $A$. This problem is similar to the standard assignment problem \cite{Kuhn}, but note that: first, $A$
is generally not a square matrix; 
second, more than one $C_i$ can be considered as new or being a 
region of occluded cells. 
We define an association list 
$\boldsymbol\zeta=(\zeta_1,\zeta_2,\dots,\zeta_{N_1})$ 
such that $\zeta_i=j$ if $C_i$ is associated with the
$j$th column of $A$. Then our problem is to find  
\begin{equation}
\label{eq:association}
\boldsymbol\zeta^*=
\underset{\boldsymbol \zeta}{\operatorname{argmax}}\; 
\sum\limits_{i=1}^{N_1} A_{i,\zeta_i}
\end{equation}
such that for any $1 \leq j \leq N_2$, there is at most one $i$ that satisfies
$\zeta_i=j$. 

\subsection{The Modified Hungarian Algorithm}
When $N_1$ and $N_2$ are both small, the derivative assignment problem
(\ref{eq:association}) can be solved using a brute-force search. 
But when $N_1$ and $N_2$ are large, 
the computational complexity of brute-force search is at least 
\begin{equation}
O\left(  \dfrac{ (\max\lbrace N_1,N_2 \rbrace ) !}
{|N_1-N_2|!}  \right) , 
\end{equation} 
which is unacceptable for real-time tracking. 
We propose a modified version of the Hungarian algorithm \cite{Kuhn}
to obtain a suboptimal solution of (\ref{eq:association}).
The computational complexity of the original Hungarian algorithm
is $O((\max\lbrace N_1,N_2 \rbrace )^3)$, 
and our modified Hungarian algorithm can achieve
a computational complexity of at most 
$O((\max\lbrace N_1,N_2 \rbrace )^3 \cdot \min\lbrace N_1,N_2 \rbrace)$. 
The modified Hungarian algorithm is given below: 
\begin{enumerate}
\item For any $1\leq i \leq N_1$, if the $i$th row of $A$ satisfies 
\begin{equation}
\underset{j}{\operatorname{argmax}}\; A_{i,j}=j^*>N_2 , 
\end{equation} 
then we associate the $i$th row with the $j^*$th column, 
and delete the $i$th row from $A$ to obtain a $N_1'\times(N_2+2)$
submatrix $A'$. 

\item \label{item:2}
Perform the standard Hungarian algorithm on the first 
$N_2$ columns of $A'$ to get an association list
$\boldsymbol \zeta$.

\item \label{item:3}
For $1\leq i \leq N_1'$, compute this value:
\begin{equation}
\Omega(A',i)=\max\lbrace A'_{i,N_2+1},A'_{i,N_2+2} \rbrace -A'_{i,\zeta_i} . 
\end{equation}
If $\max\limits_i \Omega(A',i) > 0$, let 
\begin{equation}
i^*=\underset{i}{\operatorname{argmax}}\;\Omega(A',i) .
\end{equation}
Then we associate the $i^*$th row of $A'$ with the $(N_2+1)$th column if
$A'_{i^*,N_2+1}>A'_{i^*,N_2+2}$ or the $(N_2+2)$th column if
otherwise. Now we delete the $i^*$th row from $A'$, update $N_1'$, and repeat steps
\ref{item:2} to \ref{item:3}. 

\item If $\max\limits_i \Omega(A',i) \leq 0$,
then the current $\boldsymbol \zeta$ is the final association result for
the current $A'$. 
\end{enumerate}

\subsection{Updating the List of Cells}
We update the list of cells $\lbrace L_j \rbrace$ according to the association result
of each $C_i$: 
\begin{itemize}
\item If $C_i$ is associated with a cell $L_j$ in the list ($1 \leq \zeta_i \leq N_2$),
we update the status of $L_j$ to \textit{active} and replace all its features
with the features of $C_i$.

\item If $C_i$ is considered to be a new cell ($\zeta_i=N_2+1$), we 
simply add it to the list and set its status to \textit{new}.

\item If $C_i$
is considered to be a region of overlapping cells ($\zeta_i=N_2+2$), 
we search its neighbourhood
within the radius $d_0$ for unassociated $L_j$'s that have appeared in the $(k-1)$th
frame. Then we set the status of these $L_j$'s to \textit{occluded}, and update only their 
center positions to the center position of $C_i$. 

\item For any $L_j$ that appeared in the $(k-1)$th frame and is still unassociated after all the above steps: 
if it is within the distance $d_0$ from the image border, we set its status to \textit{out}; 
otherwise, we associate it with the closest unassociated $C_i$. 
\end{itemize}

\begin{figure*}
  \centering
  \subfloat[]
  {\includegraphics[width=0.3\textwidth]{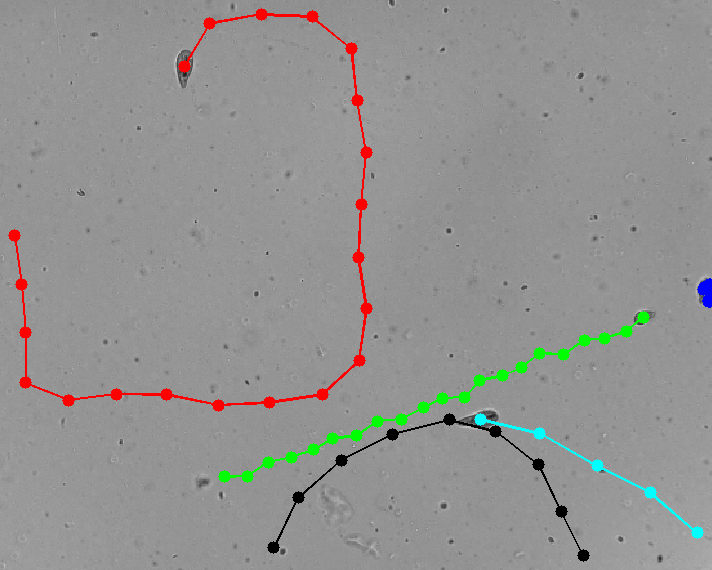} }  
  \;      
  \subfloat[]
  {\includegraphics[width=0.3\textwidth]{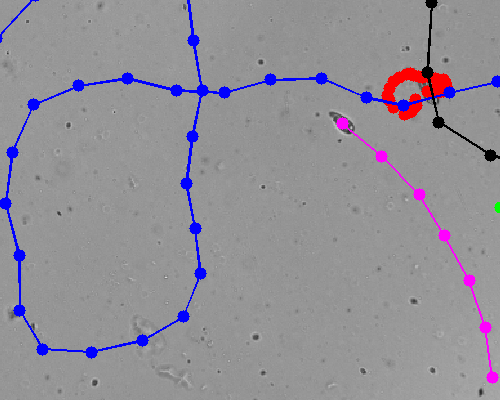} }
  \;      
  \subfloat[]
  {\includegraphics[width=0.3\textwidth]{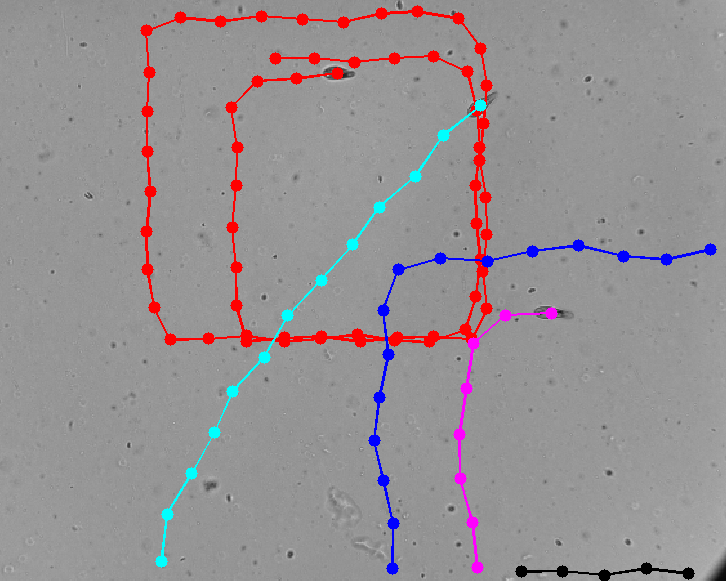} }
  \caption{Resulting cell trajectories of our method on testing videos. 
  (a) Cells getting very close. 
  (b) Two cells occluding each other (red \textit{vs.} blue, and red \textit{vs.} black).
  (c) Many cells appearing in the frame at the same time. }
  \label{fig:cell_tracking}
\end{figure*}

\section{Experiments}
To build the training set, we manually label the cell regions 
segmented in 1000 frames from 2 videos, 
where each frame is an 8-bit gray-level image of size $640 \times 512$, and 
the videos were captured at 8 frames per second \cite{cell1}. 
The typical size of a \textit{T. pyriformis} cell is between 150 and 400 pixels, 
and the typical speed is between 5 and 40 pixels per frame. 
To evaluate the performance of the decision trees $T_1$ and $T_2$, 
we repeat 20 independent experiments for different depth of the trees, 
and in each experiment 
we use randomly selected $70 \%$ of the cell pairs as the training data, 
and the rest $30 \%$ as the test data. 
The number of subdivisions of each feature is set to 1000, 
and the stop-splitting size of $\widetilde{D}$ is set to 20. 
The classifier $f$ is configured such that $f(\textbf{v})=1$ if $P_T(\textbf{v})>0.5$. 
The resulting average rates of misclassification are shown in Table \ref{table:tree_error}, 
and all 23 entries of the feature difference vector $\textbf{v}$ turn out to be useful. 

\begin{table}
\begin{center}
\begin{tabular}{ccc}
\hline
Depth of Tree &
Error Rate of $T_1$ &
Error Rate of $T_2$
\\ \hline
5 & 1.48\% & 14.02\% \\
6 & 1.46\% & 13.91\% \\
7 & 1.69\% & 12.95\% \\
8 & 1.37\% & 12.49\% \\
9 & 1.54\% & 13.07\% \\
10 & 1.55\% & 13.61\% 
\\ \hline
\end{tabular}
\end{center}
\caption{Performance of decision trees. }
\label{table:tree_error}
\end{table}

In our tracking experiments, we use decision trees of depth 8. 
The parameters $\alpha_1$, $\alpha_2$, $\alpha_3$, $\lambda_1$, $\lambda_2$
and $d_0$
are selected empirically by studying the training videos, and
the tracking results are not sensitive to minor changes of these parameters. 
Example cell trajectories obtained by our method are 
shown in Figure \ref{fig:cell_tracking}. In these examples we
set $\alpha_1=0.1$, $\alpha_2=0.1$, $\alpha_3=0.8$, 
$\lambda_1=0.00008$, $\lambda_2=0.00005$ and $d_0=70$. 

\section{Discussion and Future Work}
This paper has shown a novel cell tracking method using decision trees
as classifiers for feature difference vectors. 
The misclassification rates
of our decision trees are very low, and the tracking results of our method 
are robust against cell occlusions. Future work includes evaluating this method
on more complicated videos in which more cells exist in the frame and occlusions
of many cells 
can happen. 
Finally we will integrate our algorithm
into the real-time \textit{T. pyriformis} control system.


\bibliographystyle{latex12}
\bibliography{latex12}

\end{document}